\documentclass[10pt,twocolumn,letterpaper]{article}

\usepackage{cvpr}
\usepackage{times}
\usepackage{epsfig}
\usepackage{graphicx}
\usepackage{amsmath}
\usepackage{amssymb}
\usepackage{algorithm}
\usepackage{tabularx}
\usepackage{algpseudocode}
\usepackage{booktabs}
\usepackage{multirow}
\usepackage{subcaption}
\usepackage[pagebackref=true,hypertexnames=true]{hyperref}
\usepackage{cleveref}

\begin{document}

\title{SIG: A Synthetic Identity Generation Pipeline for Generating Evaluation Datasets for Face Recognition}

\author{Kassi Nzalasse, Rishav Raj, Eli Laird, Corey Clark \\
Intelligent Systems and Bias Examination Lab at Southern Methodist University\\
{\tt\small ejlaird@smu.edu}, {\tt\small knzalasse@smu.edu}, {\tt\small rishavr@smu.edu}, {\tt\small coreyc@smu.edu}}

\maketitle
\thispagestyle{empty}

\begin{abstract}
As Artificial Intelligence applications expand, the evaluation of models faces heightened scrutiny. Ensuring public readiness requires evaluation datasets, which differ from training data by being disjoint and ethically sourced in compliance with privacy regulations. The performance and fairness of face recognition systems depend significantly on the quality and representativeness of these evaluation datasets. This data is sometimes scraped from the internet without user's consent, causing ethical concerns that can prohibit its use without proper releases. In rare cases, data is collected in a controlled environment with consent, however, this process is time-consuming, expensive, and logistically difficult to execute. This creates a barrier for those unable to conjure the immense resources required to gather ethically sourced evaluation datasets. To address these challenges, we introduce the Synthetic Identity Generation pipeline, or SIG, that allows for the targeted creation of ethical, balanced datasets for face recognition evaluation. Our proposed and demonstrated pipeline generates high-quality images of synthetic identities with controllable pose, facial features, and demographic attributes, such as race, gender, and age. We also release an open-source evaluation dataset named ControlFace10k, consisting of 10,008 face images of 3,336 unique synthetic identities balanced across race, gender, and age, generated using the proposed SIG pipeline. We analyze ControlFace10k along with a non-synthetic BUPT dataset using state-of-the-art face recognition algorithms to demonstrate its effectiveness as an evaluation tool. This analysis highlights the dataset's characteristics and its utility in assessing algorithmic bias across different demographic groups.

\end{abstract}

\section{Introduction}

Face recognition systems are increasingly being deployed worldwide in airports \cite{tsa,chung_facial_2024,cbp,vermes_airports_2024}, sports stadiums \cite{cosentino_new_2023,gee_here_2023,cohen_facial_2023}, borders \cite{cbp2,office_facial_nodate,gutman_argemi_how_2024}, and more \cite{thales_11_2023,saravanan_facial,klosowski_facial_2020}. This growth, fueled by the rapid development of deep learning-based computer vision systems, is largely due to the growing need for fast, reliable, and fair identity verification systems in environments where security is paramount. Critical to the development and deployment of these systems is not only the data used for training, but crucially, the datasets used for evaluating the performance and fairness of face recognition models.

Ideally, evaluation face datasets should include a large number of faces of unique identities with varying features including pose, lighting, facial expressions, skin color, gender, age, and more. These datasets are essential for assessing the accuracy, robustness, and potential biases of face recognition algorithms across diverse demographic groups and conditions. However, gathering such datasets is often time-consuming, expensive, and logistically challenging.
Bias in face recognition systems has been demonstrated in most, if not all, commercial face recognition systems \cite{Howard2020QuantifyingTE,Howard2019TheEO}, and is often attributed to imbalanced datasets used for both training and evaluation \cite{Buolamwini2018GenderSI} \cite{marshall_why_2023,johnson_police}. This persistent bias issue underscores the critical need for balanced and comprehensive evaluation datasets.

The difficulties in collecting suitable evaluation data have led to problematic practices like internet scraping, raising ethical concerns and prompting new privacy legislation  \cite{drenik_data_2023}\cite{BIPA2024,CCPA2024,EUAIAct2024}. Synthetic data has emerged as a potential solution for assessing machine learning models, including face recognition systems \cite{deepface_generation,deng2020disentangled,bae2022digiface1m,qiu2021synface,kim2023dcface}, but most approaches lack fine-grained control over demographic features like as race, gender, age, and pose without specific training.

In this paper, we introduce the Synthetic Identity Generation Pipeline (SIG), designed to create balanced, ethical, and controllable evaluation datasets for face recognition systems. SIG generates high-quality images of synthetic identities with precise control over pose, facial features, and demographic attributes using crafted prompt template.

Using our SIG pipeline, we generated ControlFace10k, a new evaluation dataset for face recognition systems comprising 3,336 unique synthetic identities, balanced across race, gender, and age attributes. ControlFace10k is open-source and freely available on \href{https://huggingface.co/datasets/HuMInGameLab/ControlFace10K}{Hugging face}, for researchers to use.

Our contributions are as follows:

\begin{enumerate}
    \item Demonstration of the Synthetic Identity Generation Pipeline and its ability to generate high-quality synthetic identities with controllable pose, race, gender, and age.
    \item Release of the ControlFace10k synthetic face dataset, a demographically balanced evaluation dataset with 3,336 unique synthetic identities with varying pose.
    \item An analysis of ControlFace10k alongside a non-synthetic face dataset using similarity score distributions from state-of-the-art face recognition systems, demonstrating its effectiveness as an evaluation tool.
\end{enumerate}

\section{Related Works}

\subsection{Face Recognition Datasets}
The development and evaluation of face recognition systems are significantly dependent on the availability of face image datasets. Several prominent datasets exist, each distinguished by unique characteristics and tailored for specific applications. One of the most commonly used datasets is the Labeled Faces in the Wild (LFW) dataset \cite{lfw}, containing over $13,000$ images of public figures collected from the web. The CelebA dataset \cite{celeba}, provides more than $200,000$ celebrity images, spanning over $10,177$ identities. There is CASIA-WebFace \cite{casiawebface} which contains around half a million face images of $10,575$ real identities. VGG Face \cite{vggface} is a dataset of around $2.6$ million face images of $2622$ people. Additionally, there is MS-Celeb-1M \cite{msceleb}, which is a large scale face recognition dataset containing over 10 million face images of nearly 100,000 individuals. Notably, MS-Celeb-1M has since been withdrawn due to ethical concerns surrounding nefarious applications of face recognition and the ethical collection of data \cite{Exposing.ai}. This withdrawal underscores the critical need for carefully curated and ethically sourced evaluation datasets in the field of facial recognition and image processing.  Glint360K \cite{partialfc} is considered the ``largest and cleanest face recognition dataset'' containing over $17$ million faces of $360,232$ individuals. BUPT-Balancedface dataset \cite{bupt} contains $1.3$ million images of $28,000$ individuals divided into four race categories: African, Asian, Caucasian, and Indian. Each race group contains $7000$ individuals, making it the only dataset which balances face images based on race. \\
The LFW dataset presents many limitations that contribute to overly optimistic performance in models, including a lack of sufficient pose variance and minimal variation in age among the subjects. While attempts have been made to address these issues, they still fall short of providing a comprehensive solution. Cross-Pose LFW \cite{Zheng2018CrossPoseL}, a renovation of LFW, seeks to address the challenge of comparing faces with different poses to increase intra-class variance. However, this dataset was created through crowd-sourcing efforts to find images in LFW with larger pose differences, rather than generating controlled pose variations. This approach limits the dataset's ability to systematically evaluate pose-invariant face recognition across a wide range of controlled angles. Similarly, Cross-Age LFW \cite{Zheng2017CrossAgeLA} attempts to address the challenge of age variations, but it too is constrained by the availability of images in the original LFW dataset, lacking fine-grained control over age progression.

VGGFace2 \cite{Cao2017VGGFace2AD} offers 3.31 million face images of 9,131 individuals with variations in age, pose, illumination, ethnicity, and profession. While more comprehensive, it still suffers from several limitations that highlight the need for a more controlled approach to dataset generation. The dataset's ethnic diversity is limited by the distribution of celebrities and public figures, potentially underrepresenting certain populations. Its reliance on web-collected images introduces variations in image quality and conditions that could impact recognition performance. Moreover, the dataset exhibits a slight gender imbalance (59.3\% male) and varying numbers of images per identity (ranging from 80 to 843), which could potentially affect model training and evaluation. The web-collected nature of the images also means there is limited control over specific image conditions, potentially introducing unintended biases or variations. While efforts were made to clean the dataset, the potential for label noise remains, though it appears to be minimal.

Despite these efforts, it is important to note that there are no public face recognition datasets that comprehensively label and control for race, gender, pose, and age simultaneously. This limitation makes it challenging to evaluate face recognition algorithms across all of these attributes in a controlled and systematic manner. The shortcomings of existing datasets underscore the critical need for a more flexible, controllable, and systematic approach to dataset generation for face recognition evaluation. Such an approach would allow for precise control over demographic attributes, pose variations, and image conditions, enabling more rigorous and unbiased assessment of face recognition algorithms across diverse populations and scenarios.

\subsection{Deep Face Recognition}
The field of face recognition has seen significant advancements with the development of deep learning-based systems. However, each major contribution has had limitations that underscore the need for more robust and unbiased approaches.

DeepFace \cite{deepface} and FaceNet \cite{Schroff2015FaceNetAU} pioneered the use of deep neural networks for face recognition tasks. While groundbreaking, these early systems were limited by their reliance on large, uncontrolled datasets, potentially introducing biases and reducing generalizability across diverse populations.

SphereFace \cite{Liu2017SphereFaceDH} introduced the Angular Softmax loss, allowing CNNs to learn angularly discriminative features. Despite achieving state-of-the-art performance when trained on the CASIA dataset \cite{casiawebface}, SphereFace was constrained by the dataset's lack of demographic diversity and controlled variations in pose and age.

ArcFace \cite{Deng2018ArcFaceAA} improved upon SphereFace with its Additive Margin loss function. However, its reliance on internet-scraped datasets like CASIA \cite{casiawebface}, VGG \cite{vggface}, and MS1MV3 (a refined version of Celeb-1M \cite{msceleb}) introduced potential biases due to the uncontrolled nature of these data sources and their focus on celebrity images.

MagFace \cite{Meng2021MagFaceAU} introduced an adaptive mechanism for better within-class feature distribution. While innovative, its use of the MS1MV2 dataset \cite{Deng2018ArcFaceAA} perpetuated the limitations of internet-scraped data, lacking systematic control over demographic representation and image conditions.

Most recently, GhostFaceNet \cite{ghostface} achieved state-of-the-art performance with a lightweight model. Despite its computational efficiency, GhostFaceNet's training on MS1MV2 and MS1MV3 datasets inherited the biases and limitations of these celebrity-focused, web-scraped collections.

A critical limitation shared by all these systems is their reliance on datasets that lack sufficient diversity and balance in terms of race, gender, age, and pose. This deficiency stems from the common practice of training on internet-scraped images, which are subsequently cleaned but not systematically controlled. Consequently, these models may exhibit biases in downstream tasks, particularly when applied to diverse, real-world populations.

This persistent issue underscores the critical need for a more controlled and systematic approach to dataset generation for face recognition. Such an approach would enable the development and evaluation of face recognition systems that are more robust, unbiased, and generalizable across diverse demographics and imaging conditions.

\begin{figure}
    \centering
    \begin{subfigure}{0.5\textwidth}
        \centering
        \includegraphics[width=.75\linewidth]{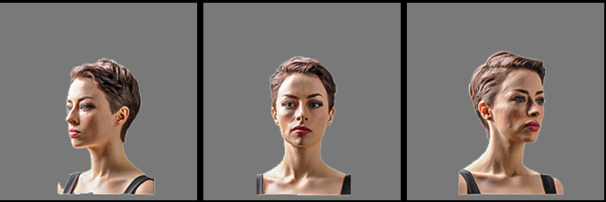} 
        \caption{Example reference image featuring diverse poses.}
        \label{fig:real-reference-image}
    \end{subfigure}\hfill
    \begin{subfigure}{0.5\textwidth}
        \centering
        \includegraphics[width=.75\linewidth]{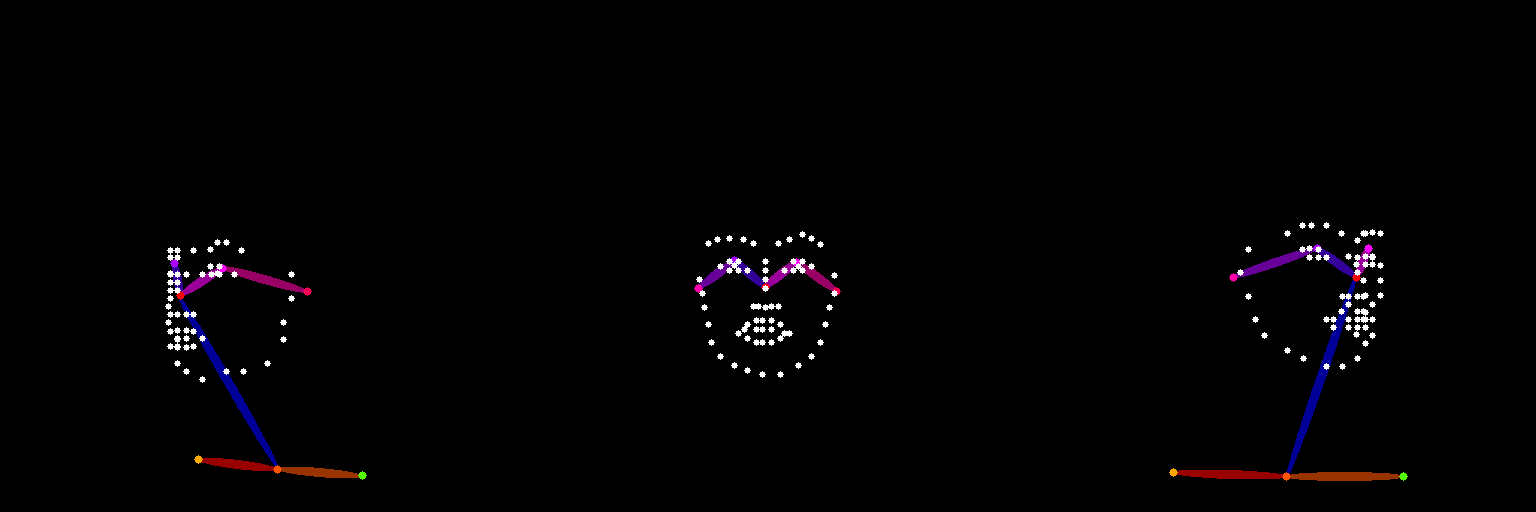} 
        \caption{Pose and facial landmarks extracted by `OpenPose' controlNet.}
        \label{fig:openpose-pose-estimation}
    \end{subfigure}
    \label{fig:main}
    \caption{(a) Example reference image featuring diverse poses, (b) Pose information extracted by the `OpenPose' ControlNet}
\end{figure}

\subsection{Synthetic Image Generation \& Datasets}

The development of synthetic generation systems has grown drastically since the introduction of Generative Adversarial Networks (GANs) in 2014. Goodfellow \etal introduced Generative Adversarial Networks (GANs) \cite{goodfellow2014generative}, which train two models simultaneously: a generator and a discriminator. GANs have found widespread application in synthetic image generation across various domains, including time-series medical records \cite{ashrafi2024protect}, image-to-image translation \cite{isola2018imagetoimage}, and more. Stable Diffusion \cite{rombach2022highresolution} uses a diffusion process to generate new images from latent space representations. It involves adding noise to the latent space iteratively, and learning to predict as well as remove that noise, which is guided by a text prompt.

There has been significant work done to build synthetic face image generators. DeepFace \cite{deepface_generation} generated face images, by firstly finetuning a pre-trained model i.e VGG-Face net, to classify facial and image characteristics, then, secondly, generating faces given some description using a custom Gaussian Mixture Model. DCGAN \cite{radford2016unsupervised} generates high-quality face images, but fails to preserve the same identity across multiple images. Microsoft's DiscoFaceGAN \cite{deng2020disentangled} aims to generate realistic face images of virtual people, with varied pose, expression, and illumination. SynFace \cite{qiu2021synface} proposes to integrate an identity mixup module to DiscoFaceGAN. It is able to generate high-quality face images but for mostly frontal-view poses. DigiFace \cite{bae2022digiface1m} introduces a large-scale synthetic dataset for face recognition training, comprising approximately 100,000 identities generated through a computer graphics pipeline. However, the dataset features synthetic images with unrealistic texture artifacts. DCFace \cite{kim2023dcface} introduces the Dual Condition Face Generator, a diffusion-based model designed to consistently generate face images of the same identity in various styles with precise control, producing 10,000 identities. However, DCFace requires the training on existing datasets for every attribute of interest, such as style, facial expression, pose, and more. Importantly, all of these methods lack mechanisms to control factors such as race, gender, age, or pose without specifically training on these features. It is crucial to highlight that there currently are no public synthetic datasets that provides comprehensive attribute labels for race, gender, pose, and age, which is one of the core reasons we have developed ControlFace10K to address this significant gap in the field.

\subsection{ControlNets}

ControlNets represent a relatively recent advancement in directing text-to-image diffusion models, particularly with Stable Diffusion, a large pre-trained diffusion model \cite{Rombach_2022_CVPR}. ControlNets enhance the control over the image generation process by freezing the original model's weights and creating a trainable copy, which receives an additional vector as input. This input helps guide the generative process, which traditionally relied solely on text prompts \cite{zhang2023adding}. A distinguishing innovation of ControlNets is their ability to generate this trainable copy without altering the foundational diffusion model's learned parameters, thereby facilitating the training and application of various ControlNets using the same unmodified base model.

The pioneering work with ControlNets has paved the way for a variety of conditional inputs to direct Stable Diffusion. The initial research incorporated a range of conditioning types, from Canny Edge \cite{4767851} and Depth Map \cite{9178977} to Normal Map \cite{vasiljevic2019diode}, M-LSD straight line detection \cite{gu2022lightweight}, HED boundaries \cite{7410521}, OpenPose \cite{cao2017realtime}, and image segmentation \cite{Zhou2017ScenePT}. This work, in particular, builds directly upon the original ControlNet implementation, utilizing the OpenPose ControlNet along with a variant of the HED boundaries, termed the LineArt ControlNet.

\section{Synthetic Identity Generation Pipeline (SIG)}
\label{sec:SIG-pipeline}

In this work, we propose the Synthetic Identity Generation (SIG) pipeline, a Stable Diffusion-based image generation pipeline. Stable Diffusion is a generative model capable of producing high-quality synthetic images from textual descriptions \cite{Rombach_2022_CVPR}. SIG leverages Absolute Reality, a finetune of Stable Diffusion, optimized for producing hyper-realistic imagery freely available on CivitAI \cite{AbsoluteReality}. The SIG pipeline utilizes this version of Stable Diffusion alongside two ControlNets to produce hyper-realistic synthetic identities with fine-grained control on identity features and pose generation.

With the implementation of meticulously crafted prompt templates, SIG is capable of generating a diverse array of unique identities, offering rich variations in demographic characteristics such as age, race, and gender.

Furthermore, the SIG pipeline is particularly effective in generating datasets for scenarios where data collection is typically challenging or even impossible. For instance, SIG can produce detailed datasets that capture a variety of human poses, enabling precise studies on posture and movement and their impact on the performance of face recognition systems—scenarios often difficult to consistently replicate in the real world due to natural human variability. Additionally, SIG can simulate facial expressions and head poses under controlled lighting conditions, as well as varying apparent ages, which are challenging to authentically capture due to the subtle nature of these factors. Unlike Generative Adversarial Networks (GANs), which rely on existing datasets to generate new data, our pipeline can synthesize identities de novo. This capability enables SIG to construct comprehensive datasets from text prompts, thereby eliminating the dependence on pre-existing imagery.

\subsection{Architecture of the SIG Pipeline}

SIG is made of two main systems: The Prompt Builder and Image Generator systems. 

\subsubsection{Prompt Builder}
\label{sec:prompt-builder}
The Prompt Builder initiates the process of creating synthetic identities - unique generated images representing individuals with consistent facial attributes across poses and lighting conditions. It creates prompts for the Image Generator system, discussed in subsequent sections, guiding the production of multiple face images per identity. These prompts include attributes such as race, background, hairstyle, facial expression, age, and gender. 

To enhance diversity, the Prompt Builder selects culturally diverse names within targeted racial demographics. We compiled 15,900 names using GPT-4 \cite{openai2024gpt4}, including 100 names (50 per gender) from 139 countries and 2,000 Indian names, evenly distributed among four racial groups. For details, see Supplementary Material.

We employ `keyword blending,' to create synthetic identities within specific demographic categories, using the syntax: $[\text{Name 1} | \text{Name 2} | \text{Name 3}]$. This technique associates each unique name triplet with a 'synthetic identity.' With 3,975 names per racial group, we can theoretically generate up to 10,460,015,075 unique name triplets per race, represented as $\binom{3975}{3}$, demonstrating the vast potential of the synthetic identity space.

To further enhance diversity, its possible to introduce more synthetic names or blend names across racial groups. For instance, increasing names per race to 5,000 could elevate possibilities to approximately 20.8 billion, showcasing our approach's scalability.

The `keyword blending' technique not only facilitates vast identity generation but also encourages the model to merge facial features associated with each name, broadening the array of unique faces produced. Our framework, as illustrated in ControlFace10K generation, provides a foundation for future research in synthetic identity generation, emphasizing flexibility and extensibility.

\begin{figure}
    \centering
    \begin{subfigure}[t]{0.48\columnwidth}  
        \centering
        \includegraphics[scale=0.15,keepaspectratio]{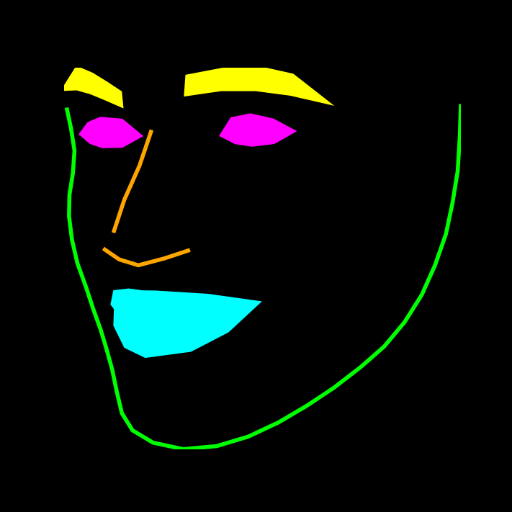}
        \caption{Conditioning Mask}
        \label{fig:single-pose-controlnet-control}
    \end{subfigure}
    \hfill
    \begin{subfigure}[t]{0.48\columnwidth}  
        \centering
        \includegraphics[scale=0.15,keepaspectratio]{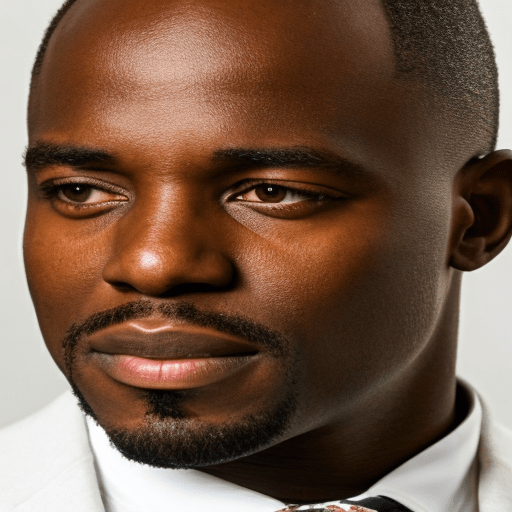}
        \caption{Generated face using conditioning mask.}
        \label{fig:single-pose-controlnet-output-african}
    \end{subfigure}
    
    \caption{Face generated by Stable Diffusion 2.1 base utilizing the trained ``pose'' ControlNet, demonstrating the model's ability to capture one head orientation}
    \label{fig:outputs-single-pose-controlnet}
\end{figure}

\subsubsection{Image Generator}
\label{sec:img-gen}
The`Image Generator' consumes prompts from the `Prompt Builder' to generate the images. The core of this component is the \textit{StableDiffusionControlNetPipeline}, from Hugging Face's Diffusers library \cite{von-platen-etal-2022-diffusers} which facilitates the integration of ControlNets \cite{Zhang2023AddingCC} within the Image Generator to guide image synthesis. For instance, the `OpenPose' ControlNet \cite{zhang2023adding} captures the desired poses for synthetic identities. It estimates pose information from a reference image featuring a person in various orientations \Cref{fig:real-reference-image}. The estimated pose information \Cref{fig:openpose-pose-estimation} is reused across generations to condition the poses featured by synthetic identities.


The Image Generator's design facilitates the integration of ControlNets, enabling face generation with controlled poses. It can operate with models hosted on Hugging Face or locally, offering deployment flexibility. Upon initialization, the Image Generator loads the previously described ControlNet and reference images. For each prompt, it uses the `OpenPose' ControlNet to capture pose information from the character reference image, then proceeds with generating the images using the available Stable Diffusion model.

\begin{figure*}[htp]
\centering
\newcommand{\person}[1]{\includegraphics[width=2.0cm]{#1}} 

\person{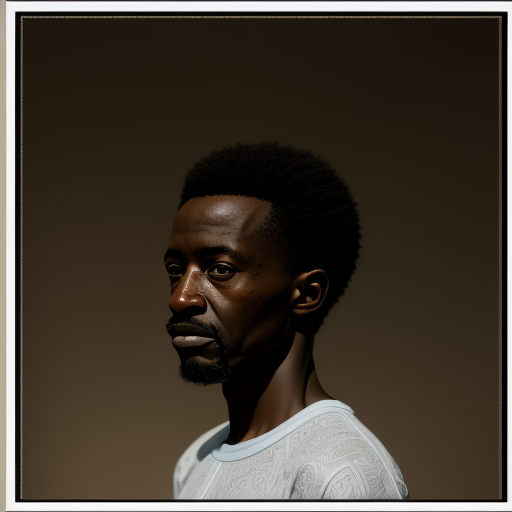}
\person{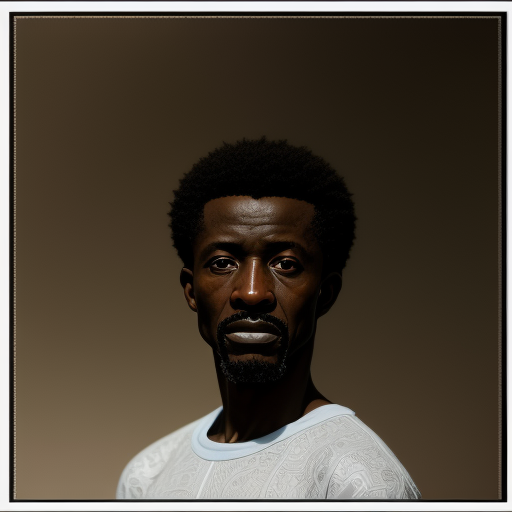}
\person{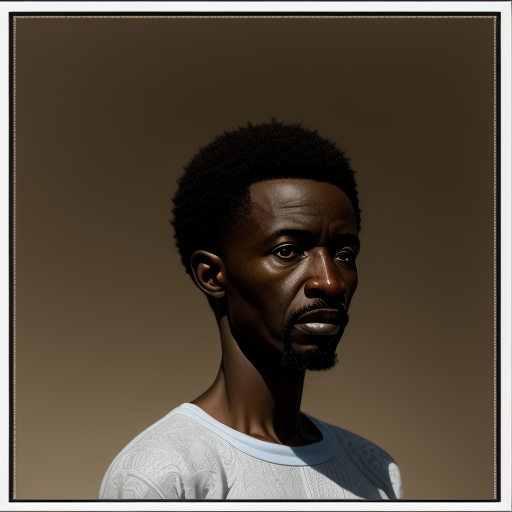}
\person{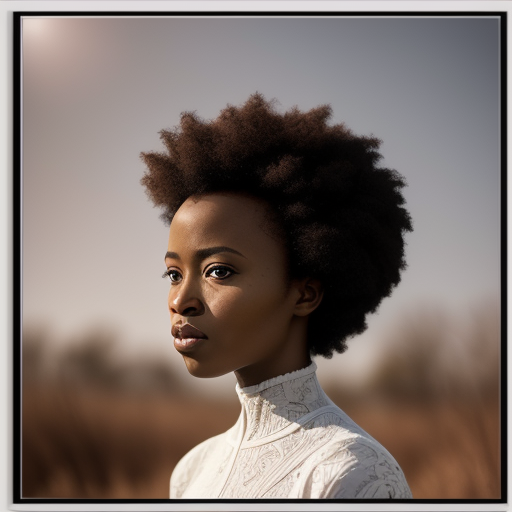}
\person{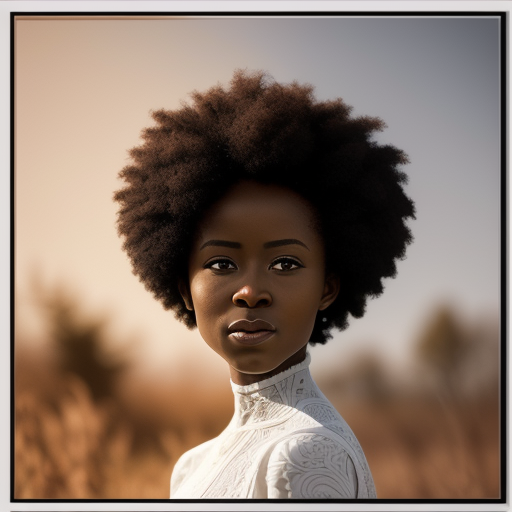}
\person{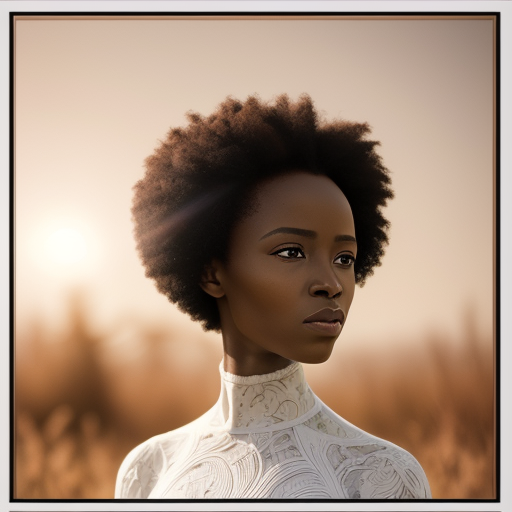}

\person{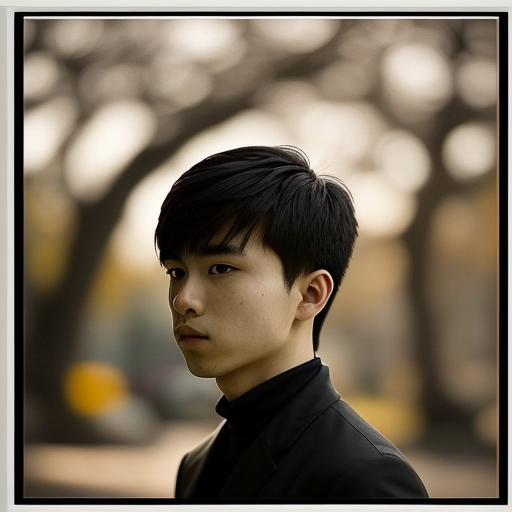}
\person{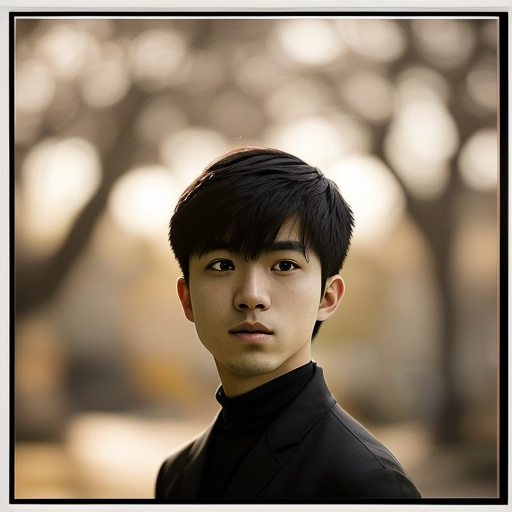}
\person{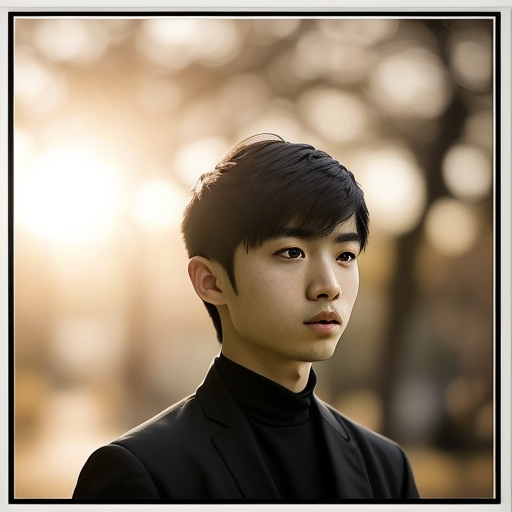}
\person{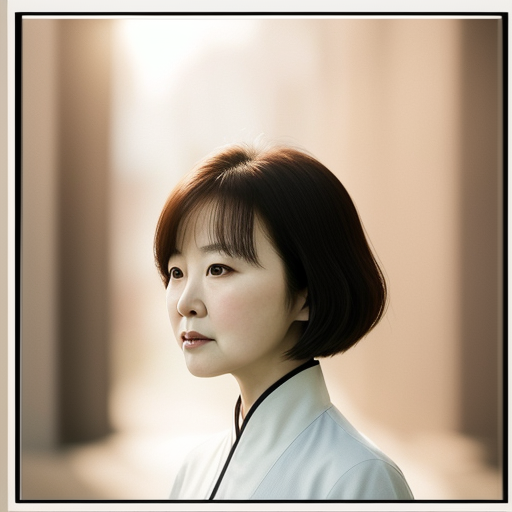}
\person{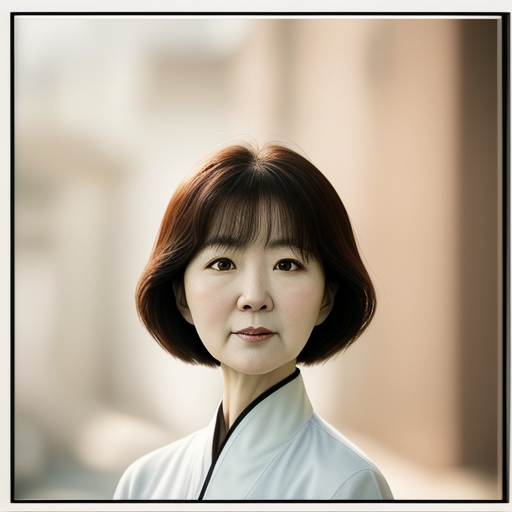}
\person{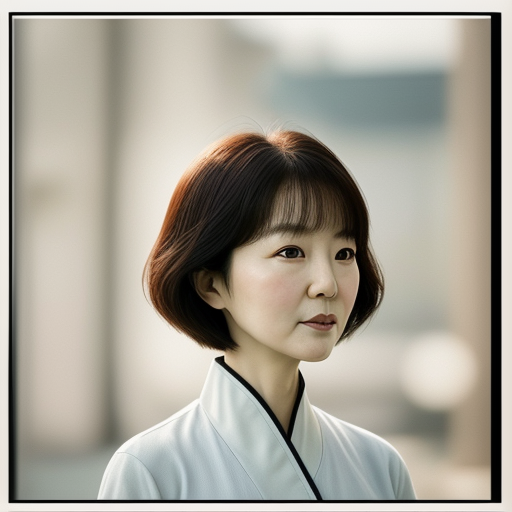}

\person{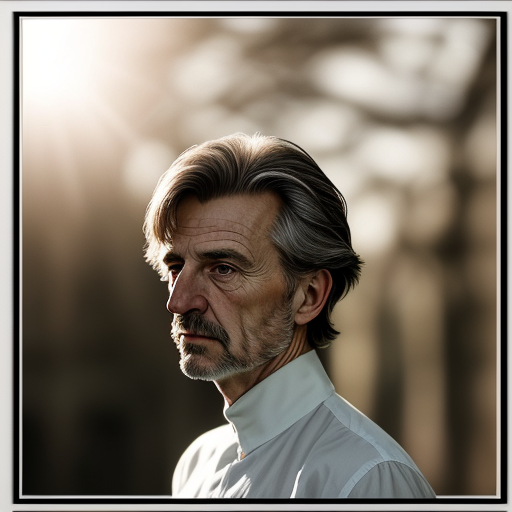}
\person{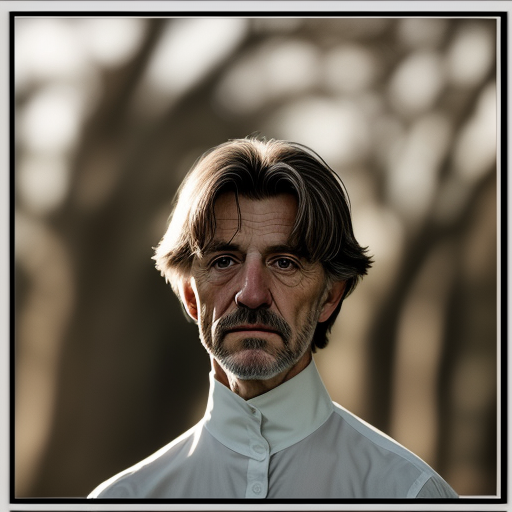}
\person{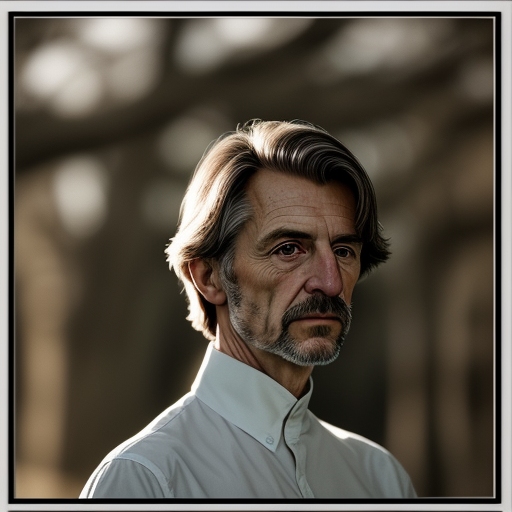}
\person{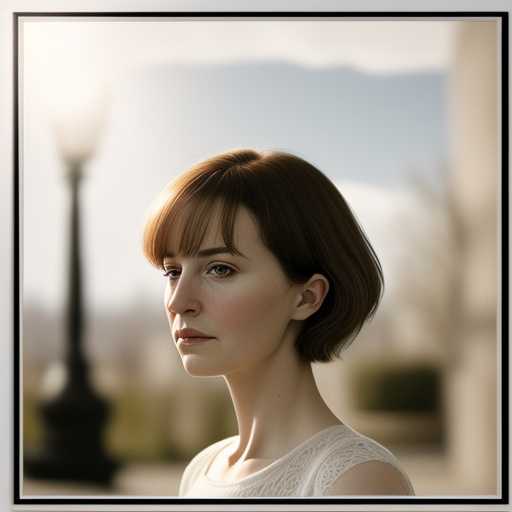}
\person{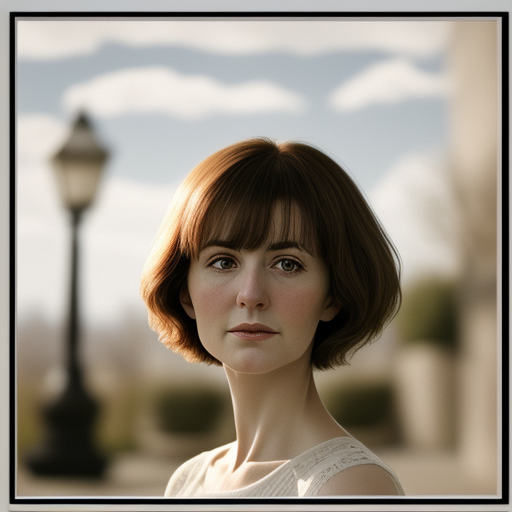}
\person{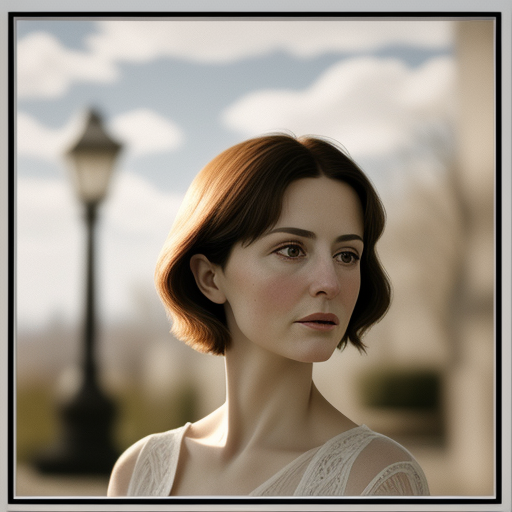}

\person{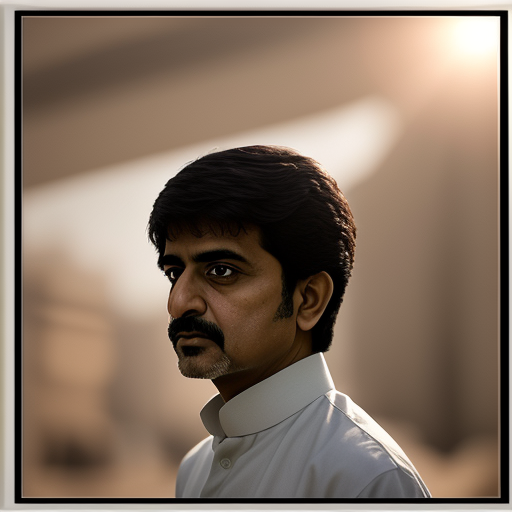}
\person{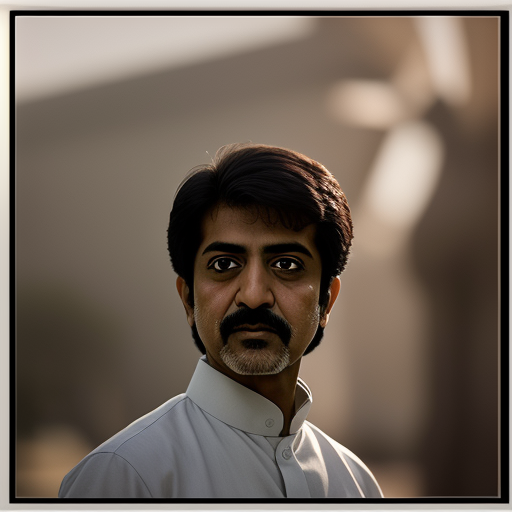}
\person{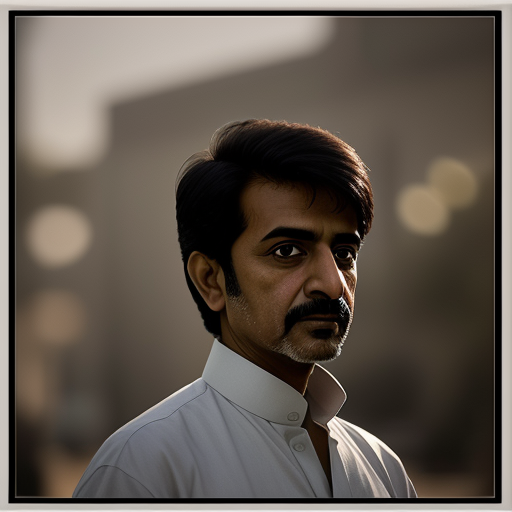}
\person{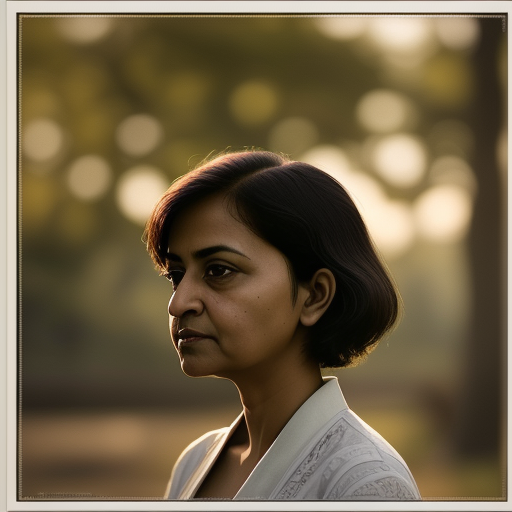}
\person{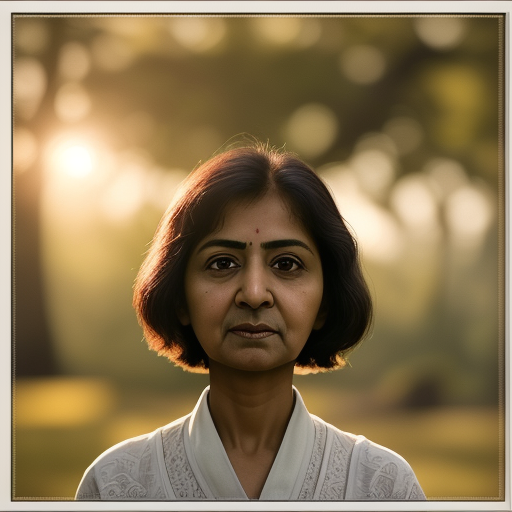}
\person{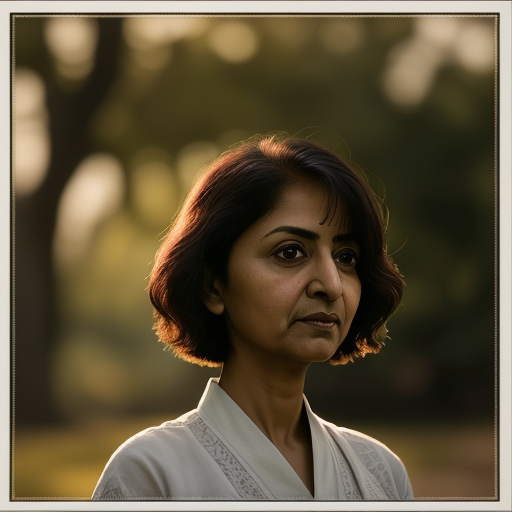}

\caption{Synthetic identities generated with SIG for each race in the ControlFace10k dataset. Each row displays images from left to right: male right-facing, male front-facing, male left-facing, female right-facing, female front-facing, female left-facing, for the races African, Asian, Caucasian, and Indian respectively. The generated identities look realistic, with no irregular textures.}
\label{fig:synthetic_identities}
\end{figure*}

\begin{table}
\centering
\begin{tabular}{@{}ccc@{}}
\toprule
\textbf{Characteristic} & \textbf{Category} & \textbf{Number of Identities} \\
\midrule
\multirow{3}{*}{\centering Age Group} & 25 & 1112 \\
                                      & 50 & 1112 \\
                                      & 65 & 1112 \\
\cmidrule{1-3}
\multirow{2}{*}{\centering Gender}    & Male   & 1668 \\
                                      & Female & 1668 \\
\cmidrule{1-3}
\multirow{4}{*}{\centering Ethnicity} & Indian   & 834  \\
                                      & Caucasian & 834  \\
                                      & African   & 834  \\
                                      & Asian     & 834  \\
\bottomrule
\end{tabular}
\vspace{.1cm}
\caption{Summary of ControlFace10k Characteristics}
\label{tab:data-breakdown}

\end{table}

\begin{figure*}[!ht]
    \centering
    \begin{subfigure}[b]{0.48\textwidth}
        \centering
        \includegraphics[width=\linewidth]{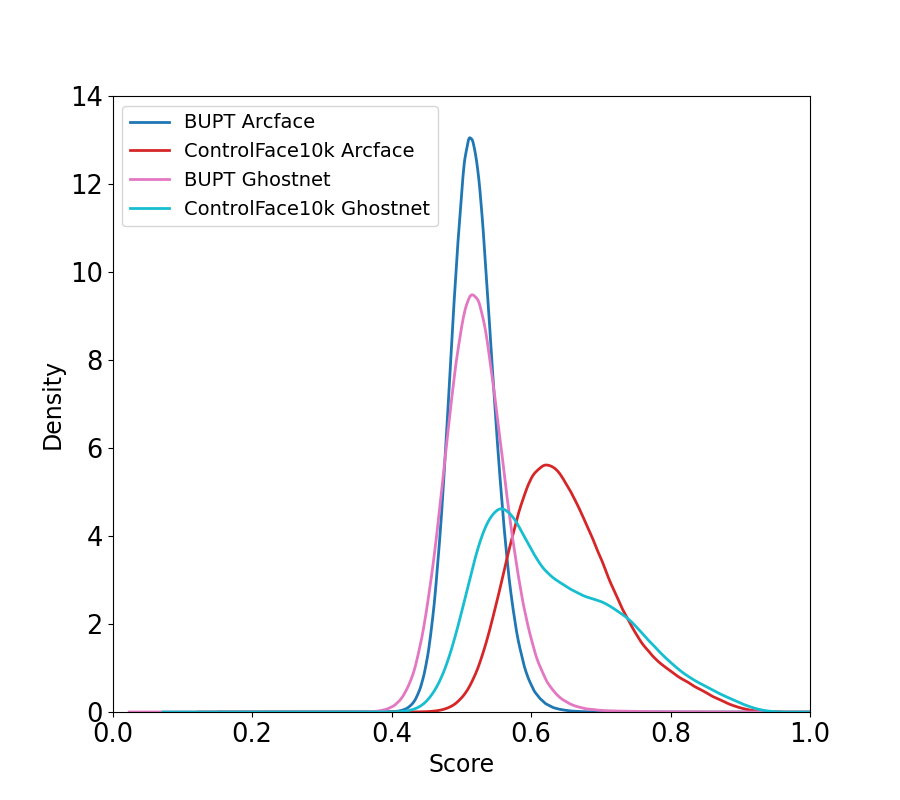}
        \caption{African}
        \label{fig:african-density}
    \end{subfigure}
    \hfill
    \begin{subfigure}[b]{0.48\textwidth}
        \centering
        \includegraphics[width=\linewidth]{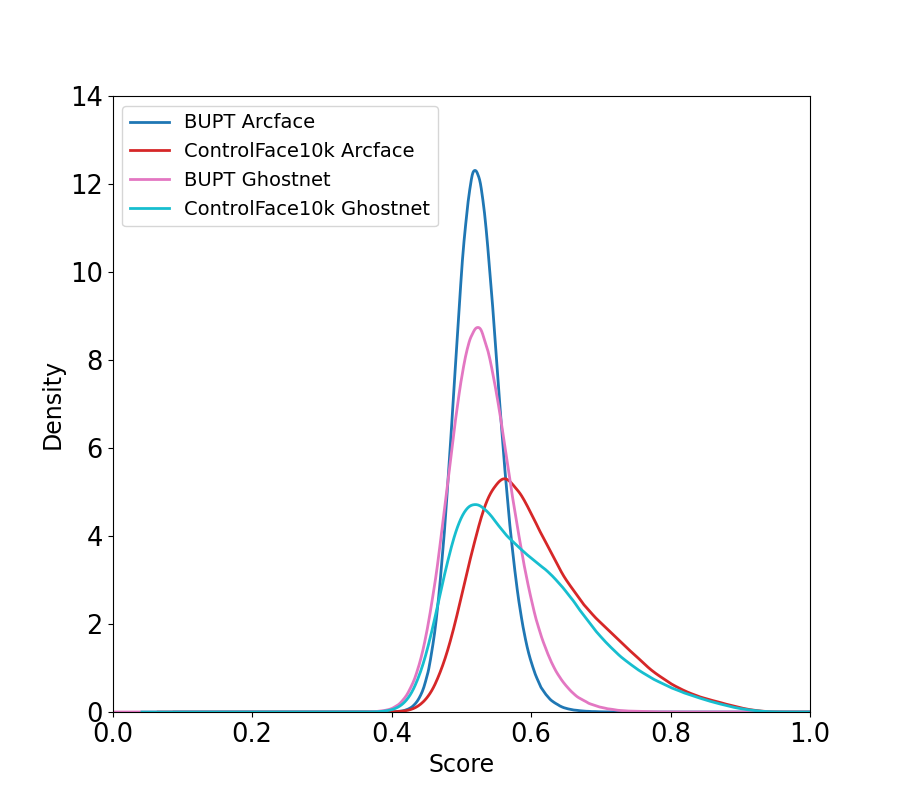}
        \caption{Asian}
        \label{fig:asian-density}
    \end{subfigure}
    
    \vspace{1em}
    
    \begin{subfigure}[b]{0.48\textwidth}
        \centering
        \includegraphics[width=\linewidth]{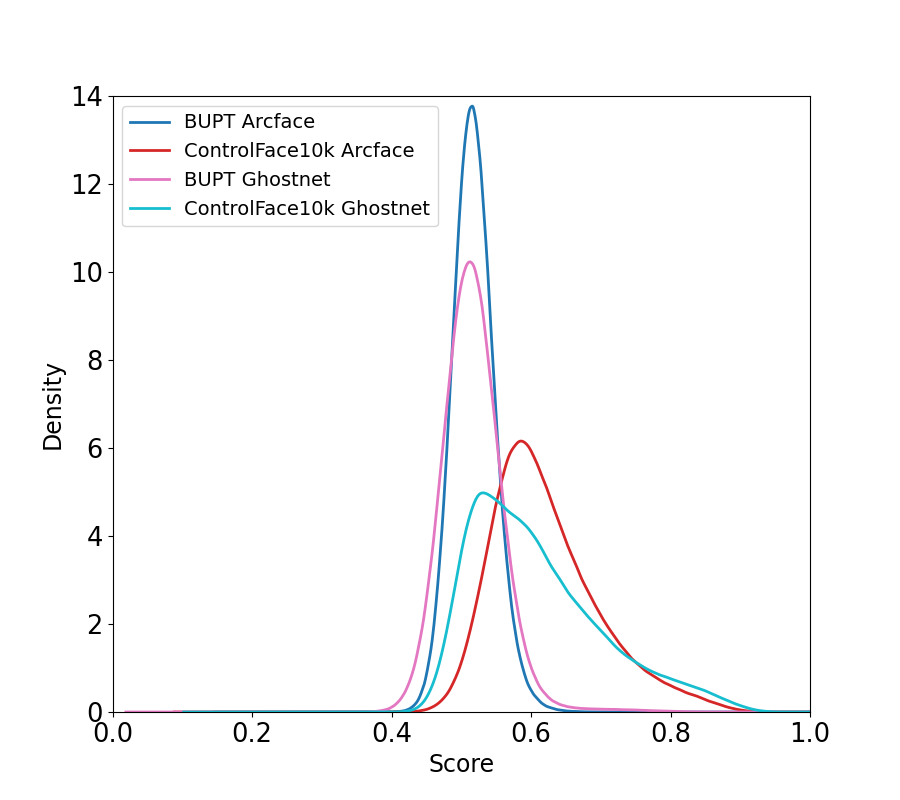}
        \caption{Caucasian}
        \label{fig:caucasian-density}
    \end{subfigure}
    \hfill
    \begin{subfigure}[b]{0.48\textwidth}
        \centering
        \includegraphics[width=\linewidth]{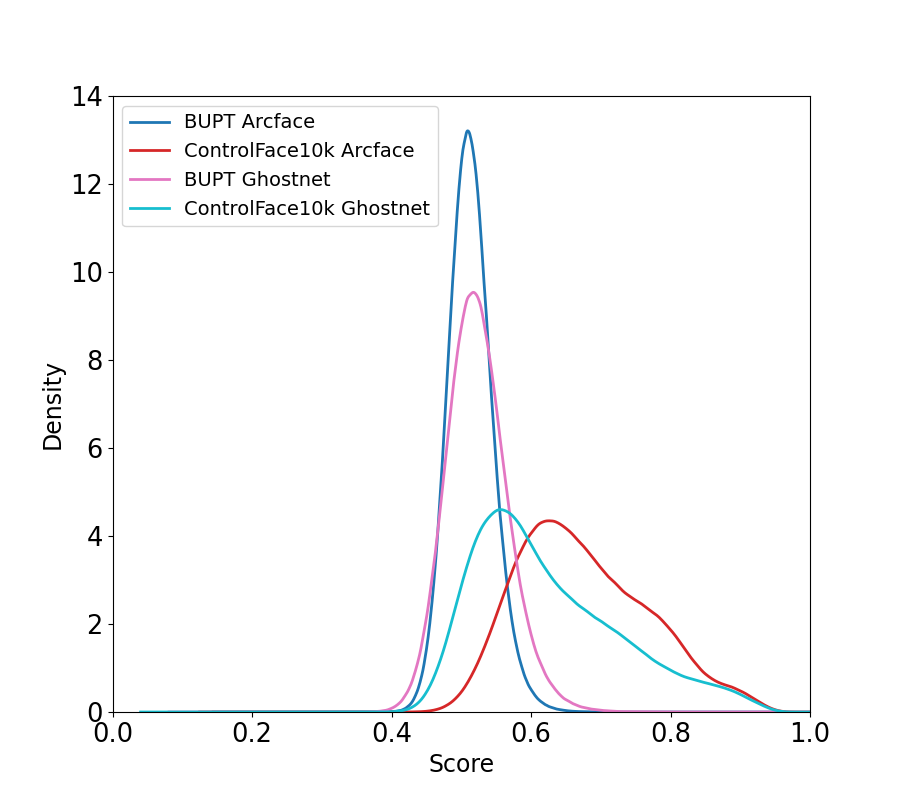}
        \caption{Indian}
        \label{fig:indian-density}
    \end{subfigure}

    \caption{Score distributions for non-mated pairs across different racial groups using ArcFace and GhostFaceNet models on BUPT and ControlFace10k datasets. Panels (a) African, (b) Asian, (c) Caucasian, (d) Indian illustrate that the ideal similarity score centered around 0.5 signifies orthogonal embeddings between different identities. The distributions typically show scores centering around 0.5, indicating effective model performance. Notably, the overlap between the synthetic data and the BUPT dataset demonstrates that the synthetic data's scores follow the behavior of non-synthetic data, which is the desired result. However, variances such as the higher scores for African identities by ArcFace compared to GhostFaceNet suggest potential biases in the algorithms. Such observations underscore the utility of synthetic data with race labels in assessing the fairness and accuracy of facial recognition technologies.}
    \label{fig:combined-density}

\end{figure*}

\section{ControlFace10k}
\label{sec:dataset}
Using the SIG pipeline described in \Cref{sec:SIG-pipeline}, we generated ControlFace10k, an evaluation face dataset containing $10,008$ images of $3,336$ synthetic identities balanced across race, gender, age and pose, \Cref{tab:data-breakdown}. This dataset is specifically designed for the evaluation and testing of face recognition systems, not for training purposes. ControlFace10k is stratified into four race groups: African, Asian, Caucasian, and Indian, with $834$ identities in each group. Each synthetic identity contains images featuring right-facing, front-facing, and left-facing poses. Synthetic identities featured in this evaluation dataset are also equally distributed across three predefined age groups of 25, 50, and 65 years. Samples of male and female synthetic identities for each race group are shown in \Cref{fig:synthetic_identities}.

It's important to note that while these synthetic images are hyper-realistic, featuring ideal lighting, focus, and contrast, our primary goal was to generate unique identities with controllable features such as pose, race, gender, and age for comprehensive evaluation. These demographic attributes are typically more challenging to generate in a controlled manner and are crucial for our evaluation purposes. Moreover, starting with high-quality images provides the flexibility to simulate lower quality scenarios through subsequent image processing, if needed.

ControlFace10k provides a balanced and controlled environment for assessing face recognition systems across various demographic groups and poses. Despite the ideal conditions of our dataset, our findings in \Cref{sec:syn-id-verification} demonstrate that models may still struggle to correctly match images across varying poses when evaluated using ControlFace10k. These discrepancies highlight the importance of having robust evaluation datasets like ControlFace10k to effectively identify and measure biases, demonstrating false positives and negatives that vary across demographic groups.

While our current focus is on controlling critical demographic attributes, it's worth noting that additional image augmentations to simulate various real-world conditions can be easily applied to the dataset. However, the primary challenge and contribution of ControlFace10k lies in its precise control over race, gender, pose, and age attributes, which are fundamental to comprehensive and unbiased evaluation of face recognition systems.

\section{Analysis}

In this section, we compare the similarity score distributions between a sample of the BUPT Balancedface dataset, referred to here as BUPT, \cite{bupt} and the ControlFace10k dataset. The BUPT sample includes 3,200 identities with an equal distribution across African, Asian, Caucasian, and Indian race groups. Our primary goal is to assess how the similarity scores for synthetic identities, generated using two state-of-the-art face recognition models, ArcFace \cite{Deng2018ArcFaceAA} and GhostFaceNet \cite{ghostface}, compare to those from real data, accounting for model performance variances. This analysis also explores intra-race comparisons among synthetic identities within the `ControlFace10k' dataset to identify potential biases introduced by the models or synthetic data generation techniques like Stable Diffusion.

We note that there exist potential overlap between BUPT and MS1MV3 \cite{ms1mv3}, the training dataset for ArcFace and GhostFaceNet, since both are independently gathered sets of 1 million celebrities. Additionally, BUPT is not controlled for pose. These factors, along with the potential familiarity of the models with some identities, may influence the similarity scores. Despite these considerations, we proceed with the analysis as BUPT remains the only freely available dataset with balanced race labels.

\subsection{Similarity Score Distributions}
\label{sec:simdist}

To understand the distribution of non-mated similarity scores in `ControlFace10k', we analyze the density curves for each race group within the dataset and compare them with the corresponding curves from the BUPT sample. This comparison provides a direct visual measure of how synthetic identities generated by each model align with real-world data. To accurately assess the similarity between images without being influenced by the different poses in the dataset, we focus on non-mated comparisons between images that are all facing forward (i.e., frontal pose).



In \Cref{fig:combined-density} we observe the Arcface and GhostFaceNet score distributions for each race group in ControlFace10k compared to BUPT. For all race groups, we notice that the BUPT scores for both models center around $0.5$. The distributions for ControlFace10k are shifted slightly to the right of the BUPT scores, but the difference is relatively small. We do note a clear difference in ControlFace10k's distributions across race groups particularly for the Arcface model. For example, in \Cref{fig:african-density} most of the mass of the Arcface distribution for African identities is above $0.6$, compared to the GhostFaceNet scores for ControlFace10k and both sets for BUPT overlapping in the $0.5$ range. A similar pattern is seen for Caucasian group though less pronounced, \Cref{fig:caucasian-density}. This shift in distributions suggests that Arcface finds these identities slightly more similar that those in the BUPT dataset. The largest shift is seen in the Arcface scores for the Indian group \Cref{fig:indian-density}, where a larger tail in the distribution suggests higher similarity for Indian identities. 

The similarity distributions computed using GhostFaceNet overlap more with the scores for BUPT than with Arcface, uncovering a `model-effect' in the distributions. This model-effect is potentially explained by GhostFaceNet's improved ability to generate distinct embeddings for synthetic identities compared to Arcface. 

Notably, both models show significant overlap with the BUPT distributions for the Asian group, \Cref{fig:asian-density}. This level of overlap for the Asian group could have two potential explanations: the Arcface and GhostFaceNet models are better at distinguishing between Asian identities, and\\or the SIG pipeline is better at generating more unique identities of Asians. Both of these explanations suggest some bias in the training data of the face recognition models and/or the Stable Diffusion model.

Overall, we see that the similarity scores produced for ControlFace10k behave similarly to the real world BUPT dataset, with a slight tail of similar identities for each race group. We also note a dependence on the face recognition model, where Arcface embeddings exhibited a slight increase in scores compared to those computed using GhostFaceNet. We also find that scores for Asian identities showed more overlap with the BUPT scores. We hypothesize that these differences in score distributions can be explained by bias in both the face recognition model and the image generation model.

\subsection{Synthetic Identity Verification}
\label{sec:syn-id-verification}
\begin{figure}
    \centering
    \includegraphics[width=\linewidth]{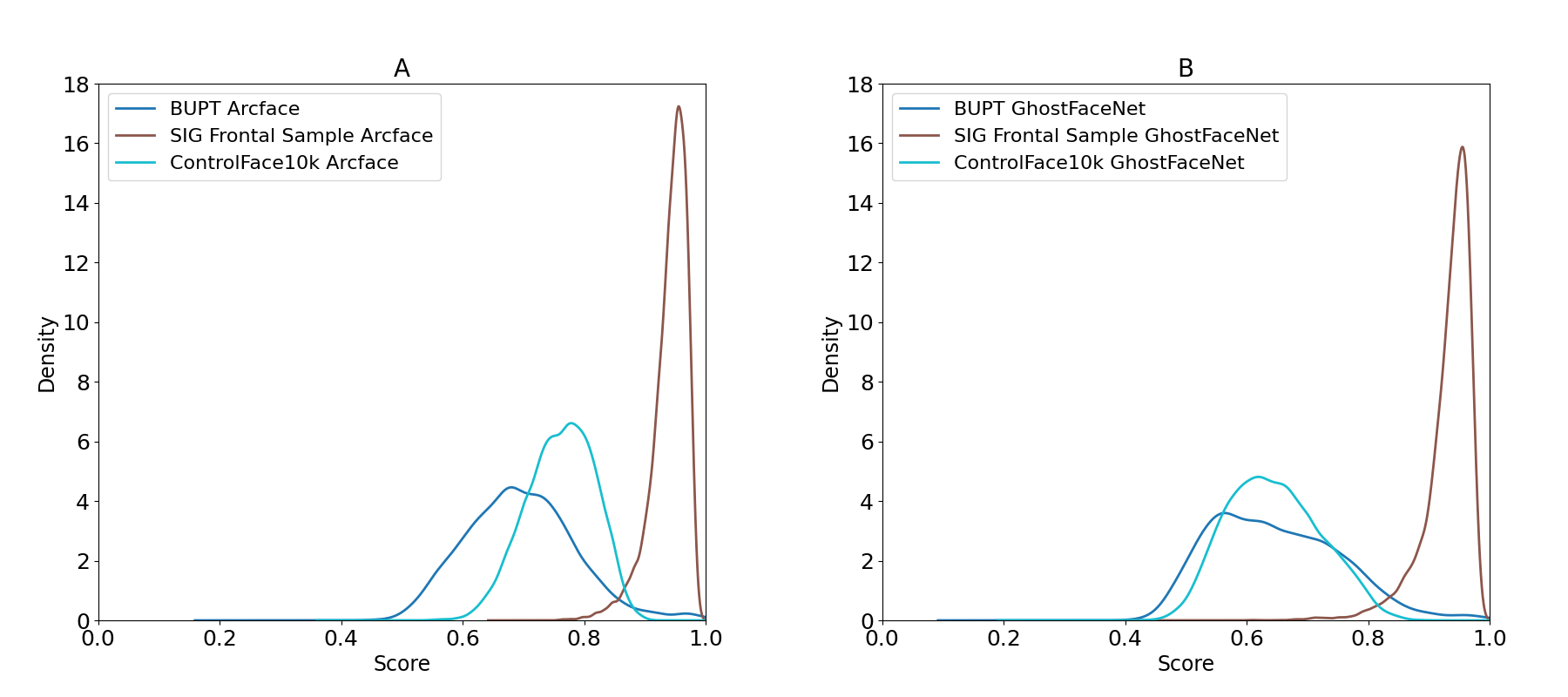}
    \caption{Facial similarity score distributions for synthetic identities using (A) ArcFace and (B) GhostFaceNet models. Each density curve represents similarity scores among three images of the same synthetic identity. Higher scores indicate greater similarity. The SIG Frontal Sample's sharp peak demonstrates high similarity when generating three frontal images per identity. In contrast, the broader distributions for BUPT (three different poses) and ControlFace10k (one frontal, one left, one right pose) datasets indicate lower similarity scores despite representing the same identity. This suggests face recognition models may be less consistent when comparing varied pose angles of the same individual.}
    \label{fig:comparison-same-identity}
\end{figure}

To demonstrate the consistency of images within synthetic identities generated by SIG, we generated an additional sample of 3360 synthetic identities, all featuring a frontal pose (SIG Frontal sample). Utilizing embedding vectors from state-of-the-art face recognition models, ArcFace and GhostFaceNet, we calculated similarity scores between \textit{distinct} images of each synthetic identity. Our goal was twofold: firstly, to validate that images belonging to the same synthetic identity maintain consistent facial attributes across different poses, as per our definition of a `synthetic identity' in \Cref{sec:prompt-builder}; and secondly, to provide an insight into the performance of current face recognition systems. We achieved this by constructing a density curve illustrating the distribution of similarity scores for synthetic identity within ControlFace10k. The resulting distributions for both models are shown in \Cref{fig:comparison-same-identity}. The comparison between the distributions of similarity scores for ControlFace10k, with varying pose, and images from the same synthetic identity, with frontal pose, indicates that while current face recognition systems perform well on frontal poses, there is room for improvement in recognizing faces across different poses. These differences can be explained by the poorer performance for both ArcFace and GhostFaceNet on non-frontal face images, as observed in their original papers \cite{Deng2018ArcFaceAA,ghostface}. 

\section{Discussion}


Synthetic data facilitates the study of specific scenarios that are difficult or expensive to capture in real-world data collection. For instance, it can enable the creation of multiple doppelganger identities across different age groups, which can be useful for examining the age-invariance of face recognition algorithms. By controlling features such as age, pose, and lighting, researchers can generate data sets that specifically target edge cases where existing datasets are sparse. This approach addresses specific challenges that might otherwise be overlooked due to the difficulties associated with collecting data in controlled environments. Synthetic data can also be used as a supplement to existing face datasets by demographically balancing them in order to reduce bias in the datasets and the training of the models. Another advantage of synthetic data lies in its capacity to uphold privacy and security standards. Synthetic data offers a solution that mimics real-world data distributions while ensuring that individual privacy is maintained. This is particularly relevant under new regulations, where the use of real data can be fraught with legal and ethical challenges.

\section{Conclusion}

In this work, we propose and demonstrate the Synthetic Identity Generation (SIG) pipeline, which generates high-quality images of synthetic identities with controllable pose, facial features, and demographic attributes,
such as race, gender, and age. Using SIG, we generated a demographically and pose balanced dataset, ControlFace10k, for use in evaluating face recognition systems. 
Through the analysis of similarity scores for the ControlFace10k dataset, we demonstrate SIG's ability to generate unique synthetic identities and discuss in detail the architecture behind our approach. To facilitate further research, we have made ControlFace10k open-source and publicly available on \href{https://huggingface.co/datasets/HuMInGameLab/ControlFace10K}{Hugging Face}. We encourage researchers and practitioners to utilize this resource for evaluating and improving face recognition systems.

Going further, we plan to scale SIG to generate much larger datasets, consisting of more unique identities and increase the controllable features in the data. We also plan to use SIG to mitigate biases in face recognition systems through targeted training of underrepresented attributes. We hope that our contributions provide new opportunities for the evaluation and development of more fair and performant face recognition systems.

{\small
\bibliographystyle{ieee}
\bibliography{mainbib}
}

\clearpage
\appendix
\section{Appendix: Supplementary Material}
\section{Average Similarity Scores by Race}

\begin{figure}
\centering
\includegraphics[width=0.48\textwidth]{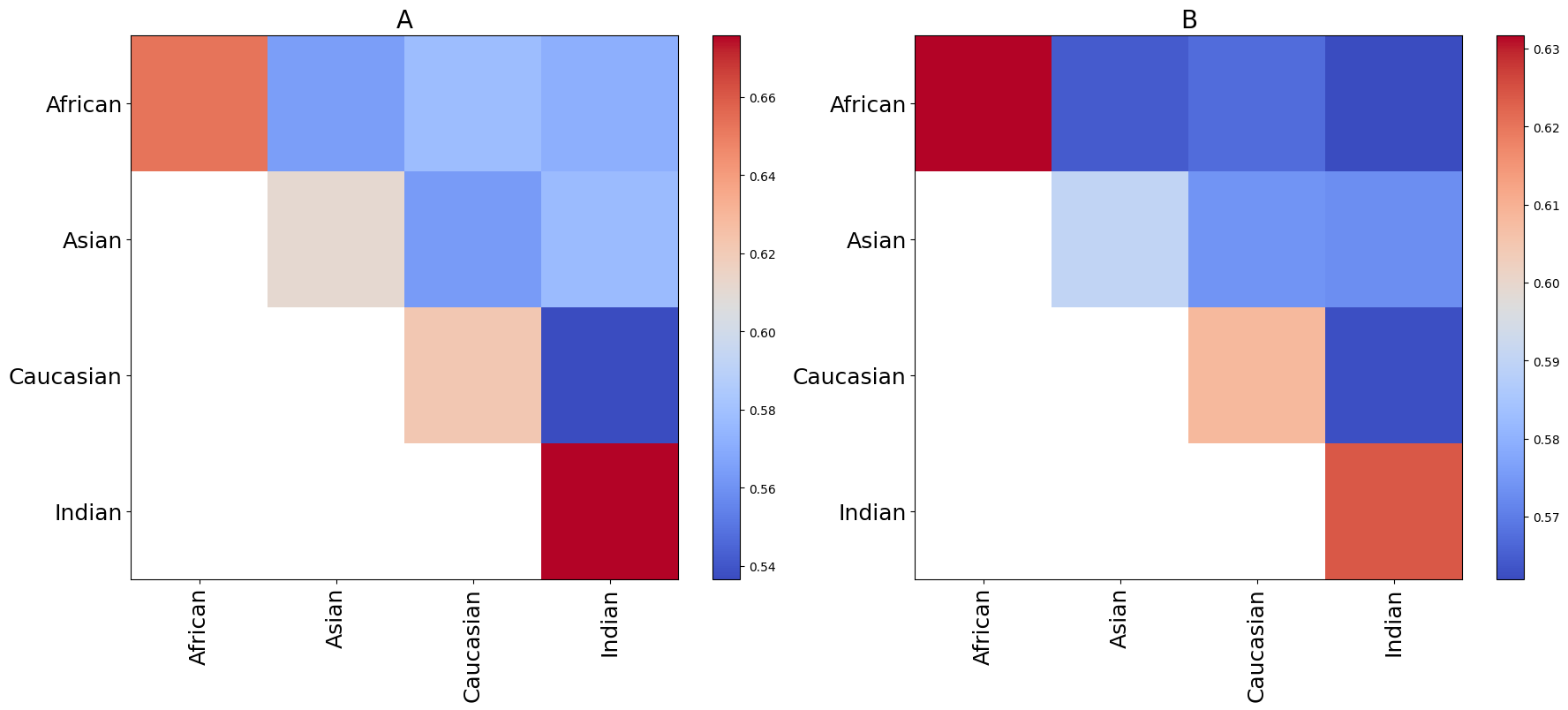}
\caption{Heatmap comparison illustrating average similarity scores by race group within the `ControlFace10k' dataset, contrasted across different facial recognition embeddings: (A) Utilizing ArcFace embeddings; (B) Utilizing GhostFaceNet embeddings}
\label{fig:avg-sim-score-arcface-ghostnet}
\vspace{-.4cm}
\end{figure}

We present the average similarity scores for synthetic identities across diverse racial groups within the ControlFace10k dataset. These scores, derived from the Arcface and GhostFaceNet models. In the provided heatmaps depicting average similarity scores from the ControlFace10k dataset, there is a noticeable contrast when comparing the outcomes of ArcFace and GhostFaceNet embeddings. It can be observed that the ArcFace-derived scores are consistently higher across all racial groups than those from GhostFaceNet, hinting at a model-intrinsic leniency in scoring identities. ArcFace scores indicate a closer resemblance within synthetic identities of the Indian group, while GhostFaceNet scores imply a similar closeness among African identities. These patterns may stem from the models' differential sensitivity to skin tone variations, which range from very dark to brown in African identities and from fair to dark brown in Indian identities.

\section{Countries Used to Generate Names for Each Race}

SIG's Prompt Builder module generates a prompt for each desired synthetic identity. To foster the uniqueness of these prompts and enhance the diversity of the resulting portraits, the builder selects names that reflect diverse cultural or linguistic backgrounds within the targeted racial demographic. These names, synthetically generated, represent common names from specific racial groups. To compile these names, we initially identify countries where the racial group is predominantly found.

The GPT-4 large language model is then utilized to produce names that are found in the identified countries, see \Cref{tab:countries_by_region} below, with a target of 50 names per gender per country. From this effort, we gathered 100 names (50 per gender) from 139 countries, plus an additional 2,000 names specifically for Indian demographics, resulting in a total of 15,900 names. These are evenly distributed among four racial groups, providing approximately 3,975 names per group. 

\begin{table}[!h]
\centering
\caption{Countries used to generate names for each race.}
\label{tab:countries_by_region}
\begin{tabular}{@{}llll@{}}
\toprule
African & Asian & Caucasian & Indian \\ \midrule
Algeria & Afghanistan & United States & India \\
Angola & Armenia & Canada &  \\
Benin & Azerbaijan & United Kingdom &  \\
Botswana & Bahrain & France &  \\
Burkina Faso & Bangladesh & Germany &  \\
Burundi & Bhutan & Italy &  \\
Cabo Verde & Brunei & Spain &  \\
Cameroon & Cambodia & Russia &  \\
Central African Republic & China & Australia &  \\
Chad & Cyprus & New Zealand &  \\
Comoros & Georgia & Sweden &  \\
Democratic Republic of the Congo & India & Norway &  \\
Republic of the Congo & Indonesia & Denmark &  \\
Djibouti & Iran & Finland &  \\
Egypt & Iraq & Netherlands &  \\
Equatorial Guinea & Israel & Belgium &  \\
Eritrea & Japan & Austria &  \\
Eswatini & Jordan & Switzerland &  \\
Ethiopia & Kazakhstan & Ireland &  \\
Gabon & Kuwait & Portugal &  \\
Gambia & Kyrgyzstan & Iceland &  \\
Ghana & Laos & Greece &  \\
Guinea & Lebanon & Poland &  \\
Guinea-Bissau & Malaysia & Czech Republic &  \\
Ivory Coast & Maldives & Hungary &  \\
Kenya & Mongolia & Slovakia &  \\
Lesotho & Myanmar & Croatia &  \\
Liberia & Nepal & Slovenia &  \\
Libya & North Korea & Bulgaria &  \\
Madagascar & Oman & Romania &  \\
Malawi & Pakistan & Estonia &  \\
Mali & Palestine & Latvia &  \\
Mauritania & Philippines & Lithuania &  \\
Mauritius & Qatar & Luxembourg &  \\
Morocco & Russia & Malta &  \\
Mozambique & Saudi Arabia &  &  \\
Namibia & Singapore &  &  \\
Niger & South Korea &  &  \\
Nigeria & Sri Lanka &  &  \\
Rwanda & Syria &  &  \\
São Tomé and Príncipe & Taiwan &  &  \\
Senegal & Tajikistan &  &  \\
Seychelles & Thailand &  &  \\
Sierra Leone & Timor-Leste &  &  \\
Somalia & Turkey &  &  \\
South Africa & Turkmenistan &  &  \\
South Sudan & United Arab Emirates &  &  \\
Sudan & Uzbekistan &  &  \\
Tanzania & Vietnam &  &  \\
Togo & Yemen &  &  \\
Tunisia &  &  &  \\
Uganda &  &  &  \\
Zambia &  &  &  \\
Zimbabwe &  &  &  \\ \bottomrule
\end{tabular}
\end{table}



\end{document}